# 1000× Faster Camera and Machine Vision with Ordinary Devices

Tiejun Huang[a], Yajing Zheng[a], Zhaofei Yu[a]*, Rui Chen[a], Yuan Li[a], Ruiqin Xiong[a], Lei Ma[a], Junwei Zhao[a], Siwei Dong[a], Lin Zhu[a], Jianing Li[a], Shanshan Jia[a], Yihua Fu[a], Boxin Shi[a], Si Wu[a] & Yonghong Tian[a] a Department of Computer Science and Technology, National Engineering Laboratory for Video Technology, Peking University, Beijing, China.

## Abstract

In digital cameras, we find a major limitation: the image and video form inherited from a film camera obstructs it from capturing the rapidly changing photonic world. Here, we present vidar, a bit sequence array where each bit represents whether the accumulation of photons has reached a threshold, to record and reconstruct the scene radiance at any moment. By employing only consumer-level CMOS sensors and integrated circuits, we have developed a vidar camera that is 1,000× faster than conventional cameras. By treating vidar as spike trains in biological vision, we have further developed a spiking neural network-based machine vision system that combines the speed of the machine and the mechanism of biological vision, achieving high-speed object detection and tracking 1,000× faster than human vision. We demonstrate the utility of the vidar camera and the super vision system in an assistant referee and target pointing system. Our study is expected to fundamentally revolutionize the image and video concepts and related industries, including photography, movies, and visual media, and to unseal a new spiking neural network-enabled speed-free machine vision era.

*E-mail address*:yuzf12@pku.edu.cn

# 1. Introduction

Is a digital camera truly digital? The typical answer is yes, as imaging on film is replaced by imaging with charge-coupled device/complementary metal-oxide-semiconductor (CCD/CMOS) sensors and digital circuits. However, the essence of the digital camera remains in the analog age, as it indiscriminately inherits the image and video form, which are necessary to record the temporal dynamics of light on film [1-3] but are not necessary for a purely digital system. In fact, an image cannot record the change in light during the exposure time, and a video even misses all of the dynamic information between two neighboring exposures. Furthermore, the frame rate of cost-effective consumer-level cameras is only tens of hertz, making capture of high-speed scenes impossible. In contrast, high-speed cameras can reach a time sampling frequency of thousands or even tens of thousands of hertz, but they require specialized sensors and shutters that are highly expensive [4-5]. Therefore, the image and video form has become the greatest obstacle for digital cameras to capture the fast-changing photonic world.

In this study, we propose a revolutionary visual representation, called vidar, that breaks the conventional frame-based representation,

allowing cost-effective high-speed cameras to be made. Inspired by the sampling mechanism of primate fovea [6-7], vidar takes advantage of spike sequences to represent the changes in light in the spatial-temporal domain and can accurately retain the timing of physical optical flow. This brings the ability to reconstruct the scene radiance at any given moment, which is called full-time imaging.

Based on the new visual representation model, we develop the VidarOne chip and vidar camera with the same CMOS sensors and consumer-grade integrated circuits as in traditional cameras [8]. A spike is generated when the accumulated intensity collected by the photosensitive devices exceeds a given threshold. The photoelectric conversion speed of these photosensitive devices is approximately 10 nanoseconds, which is six orders of magnitude faster than that of a human retina [9]. Therefore, while having similar mechanisms, the vidar camera avoids the speed limitation of biological vision. The first vidar camera we developed has a time sampling frequency of 40,000 Hz, which can be used to implement high-speed imaging 1000 times faster than that of human vision and conventional cameras.

The spike streams generated by the vidar camera have a clear physical meaning; that is, they encode the spatial-temporal visual information of the input scene and can thus be used to perform high-speed vision tasks. However, conventional machine vision methods based on artificial neural networks (ANNs) [10] cannot process these spike streams in real time because they must first convert the spike streams to images (40,000 frames/second) and then process them frame by frame. In contrast, we find that spiking neural networks (SNNs) [11-12] can naturally process the output spike streams of a vidar camera in real time. Using this approach, we developed an SNN-based supervision system that combines the speed of the machine and the mechanism of biological vision [13-18]. The vision process can be understood as the flow of spike sequences within the SNN; thus, the processing speed only depends on the physical properties of the SNN. We realized real-time processing of 40,000 Hz vidar spike streams in the supervision system with ordinary central processing units (CPUs) and achieved high-speed moving object detection and tracking that is 1000 times faster than that of human vision. In the future, by using SNN hardware and higher speed vidar cameras, we can implement object detection, tracking, prediction, and recognition at electrical speeds and achieve superhuman vision faster by more orders of magnitude. All we need are regular consumer-grade optoelectronic devices and circuit technologies that are widely used today.

## 2. Methods

**Visual texture reconstruction.** The TFW method obtains the pixel value (proportional to the scene radiance) by calculating the number of spikes in a time window. Specifically, a moving time window collects spikes in a specific period. By counting these spikes, the pixel value is estimated by:

$$P_{t_i} = \frac{N_w}{w} \cdot C, \tag{1}$$

where $P_{t_i}$ refers to the pixel value at moment $t_i$, $w$ is the size of the time window that contains the previous $w$ moments before $t_i$, $N_w$ is the total number of spikes collected in the time window, and $C$ refers to the maximum dynamic range of the reconstruction. The TFI method assumes that the scene radiance $\bar{I}$ is a constant in a short period. According to the vidar camera mechanism, the spike generation condition can be simplified as $\bar{I}\Delta t \geq \phi$, where $\Delta t$ is the interspike interval obtained by calculating the time between two neighboring spikes and $\phi$ denotes the trigger threshold. Thus, the pixel value can be estimated with two spikes (i.e., one interspike interval):

$$P_{t_i} = \frac{C}{\Delta t_i}, \qquad (2)$$

where $\Delta t_i$ represents the interspike interval corresponding to moment $t_i$.

We test the proposed image reconstruction algorithms and compare it with conventional camera. We build a hybrid camera system consisting of the Vidar camera, conventional camera and a beam splitter. Two cameras can record the same scene through the beam splitter. we employ two no-reference image quality assessment metrics, namely two-dimensional (2-D) entropy and standard deviation (STD). 2-D entropy uses both the gray value of a pixel and its local average gray value to evaluate the amount of information carried by the image, larger 2-D entropy means more information. STD evaluates the contrast of the image, and larger STD means higher contrast. As shown in Table 1, our reconstruction methods achieve better results than conventional camera in all two metrics.

**Table 1. Comparison among TFI, TFW and conventional camera**

| | | TFI | TFW | conventional camera |
|---|---|---|---|---|
| STD | motion | 73.81 | **74.33** | 73.25 |
| | static | 73.82 | **74.29** | 73.44 |
| 2D-Ent | motion | 12.86 | **13.17** | 11.54 |
| | static | 12.83 | **13.21** | 11.78 |

**Dynamic connection gate.** The dynamic connection gate is based on short-term plasticity (STP), which refers to the short-term change in synaptic strength (usually between tens to thousands of milliseconds), also known as the dynamic connection between neurons [29-20]. When a postsynaptic neuron receives a sequence of action potentials from a presynaptic neuron, the postsynaptic potential (PSP) changes according to:

$$\text{PSP}(t) = A \cdot x(t) \cdot u(t), \qquad (3)$$

where $A$ is the maximum current value that an action potential can trigger on a postsynaptic neuron, $x(t)(0 < x(t) < 1)$ represents the remaining number of available neurotransmitters in the axon terminal at time $t$, and $u(t)$ denotes the release probability of neurotransmitters in the axon at time $t$. When a postsynaptic neuron receives a sequence of action potentials with fixed frequency from a presynaptic neuron, the PSP converges to a stable state after several spikes arrive [21] (Fig. 1a).

If the spike frequency changes, then the PSP will fluctuate around a stable value (Fig. 1b). By taking advantage of the sensitivity of the STP to the release time mode of the input spike streams, the spike streams generated by the background or static areas can be filtered, and only the spike streams generated by the moving object are retained.

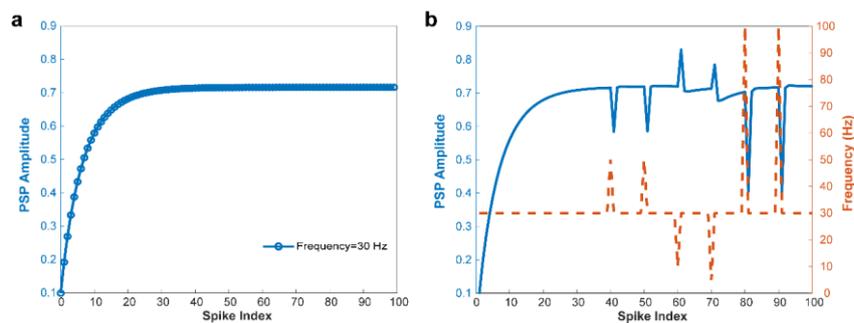

Fig. 1. PSP change with respect to the spikes from a presynaptic neuron. **a,** The PSP amplitude (blue curve) converges to a stable state for the input spikes with a fixed frequency of 30 Hz. **b,** The PSP amplitude (blue curve) fluctuates around a stable value for the input spikes with varying frequency (red curve).

**Detection and tracking**. The neuron in the filter layer is connected to 9 adjacent LIF neurons in the detection layer. Each LIF neuron accumulates current from presynaptic neurons and fires when the membrane potential reaches the threshold. As only the area corresponding to a moving object in the detection layer can generate spike streams, each moving object can be found by detecting the connected area of the firing neurons. The tracking-by-detection method is utilized to track different moving objects. To evaluate algorithm accuracy, we use the detection success rate (DSP) to measure the effect of object detection, and MOTA (multiple-object tracking accuracies), FP (false positive), FN (miss detected), IDS (ID switches) to evaluate the effect of object tracking. The results are shown in Table 2, we can find that our algorithm can achieve good performance with low power.

**Table 2. Accuracy of detection and tracking**

| DSP | MOTA | FP | FN | IDS | Speed | Power |
|---|---|---|---|---|---|---|
| 100% | 96.32% | 23 | 23 | 0 | 20811Hz | 2.254W |

**CANN for prediction.** A CANN is a canonical network model for neural information representation. A previous study has revealed that by adding negative feedback to neuronal dynamics, a CANN can track a moving object anticipatively with an approximately constant leading time [22]. Based on this, we develop a CANN model for anticipatively tracking a fast-moving object in real-world applications, with the visual inputs coming directly from the vidar camera.

**Object recognition.** The synaptic weights of the recognition SNN are trained with the BP-STDP learning rule, which is derived from Tavanaei and Maidi [23]. Here, we use multiple spike neurons to represent one category in the last layer (Fig. 5e). Specifically,

$m$ neurons are divided into $n$ groups to represent $n$ categories ($m=kn$). The real spike data generated from the vidar camera mixed with the simulated data generated from Vidar-Sim are used as the training dataset. In the training process, the classification neuron with the maximum membrane potential in the target group updates its synaptic weights according to STDP [24], while the misfired classification neurons with the maximum membrane potential in the nontarget groups undergo anti-STDP [25]. Then, the modulation of synaptic weights is backpropagated layer by layer according to presynaptic activities.

**Estimation of velocity.** As the angular velocity of the fan is 2400 revolutions per minute (rpm) and the distance from the center of the characters to the center of the fan is 0.12 m, the linear velocity of the characters on the fan can be estimated as $2400/60 \times 2 \times \pi \times 0.12 \approx 30$ m/s. Considering that the distance between the fan and the vidar camera is 0.75 m, the vidar camera and the super vision system can detect, track and recognize a moving object with a linear velocity of 40 m/s within 1 m in real time according to the central perspective principle.

**Vidar high-speed spike dataset (VHSSD)**. This dataset includes (1) spike streams of high-speed moving targets captured with a static vidar camera (Class A) and (2) spike streams of natural scenes captured with a high-speed moving vidar camera (Class B). Class A contains a moving car, a rotating disc, a rotating fan, and a bursting balloon, while Class B contains train, forest, viaduct bridge, and railway scenes (more details can be found in Table 3). We also provide VidarPlayer for playback of the spike sequences.

Table 3. Unified description of the VHSSD

| Sequence | | Length (s) | Spike number |
|---|---|---|---|
| Class A: moving target | Moving Car (100 km/h) | 0.2 | 102206031 |
| | Rotating Disc (7200 rpm) | 3.84 | 535852602 |
| | Rotating Fan (2400 rpm) | 2 | 407620564 |
| | Bursting Balloon | 0.1 | 6351184 |
| Class B: moving vidar camera | Moving Train (350 km/h) | 0.2 | 42898223 |
| | Forest | 0.22 | 93319068 |
| | Viaduct Bridge | 0.22 | 136859111 |
| | Railway | 0.22 | 87866720 |

**VidarPlayer.** This visualization software can play real and simulated spatial-temporal spike streams (i.e.dat files) recorded by the vidar camera, providing high frame rate videos reconstructed with the proposed TFW and TFI. VidarPlayer supports various resolutions, such as 400×250, and even extends the simulator's compatibility.

**Vidar-Sim.** Vidar-Sim is a simulator of the vidar camera used to simulate arbitrary camera motion and object motion in 3D scenes and provides reference images and additional information, including camera pose, object velocity, etc. This simulator integrates the principle of vidar camera theory and multiple rendering engines, including a fast, custom renderer developed based on OpenGL that can render and generate spike streams in real time and a photorealistic render based on Blender's Cycles engine.

## 3. Results

*3.1 Vidar: a new and more natural visual form*

Before we introduce the new visual representation called vidar, we briefly review the concepts of images and videos. For a large number of photons traveling within a camera's viewing frustum (Fig. 2a), the camera will acquire images at time $t_1, t_2, t_3, ...$ (the time interval is $\frac{1}{f}$ seconds) according to the predetermined frame rate $f$. During image capturing, all the photosensitive units simultaneously capture photons over a duration of $\Delta t$ (known as the exposure time, $\Delta t < \frac{1}{f}$) and then record the accumulated intensities. The distribution of intensities according to the spatial arrangement of the photoreceptors forms the image, and a sequence of images arranged at equal time intervals is called a video. From the perspective of the plenoptic function [26], what is recorded in the image is not the moment at time $t_1, t_2, t_3, ...$ but the accumulation of physical processes that last $\Delta t$. For a video, the cumulative time (exposure time) $\Delta t$ used to acquire each frame is less than or equal to the time interval $\frac{1}{f}$ between two frames of the video, which means that the information in period $\frac{1}{f} - \Delta t$ is completely lost, and the motion process during time interval $\Delta t$ is also 'squashed' into the image and lost. Thus, the temporal domain sampling of the video is not a complete sampling of the physical process.

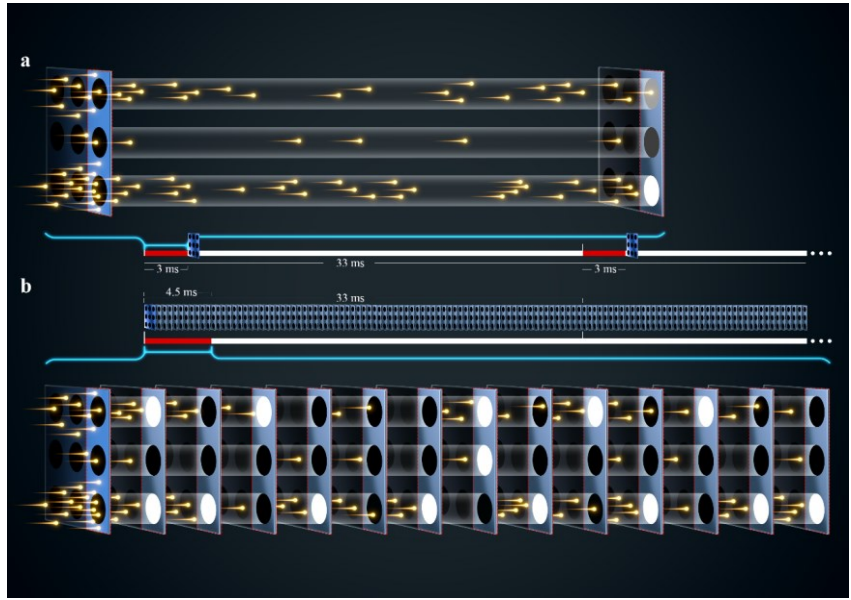

**Fig. 2. Overall comparison of image and vidar in terms of visual information representation. a,** Visual representation by images and video. The photosensitive units (three circles) capture a group of photons (yellow shooting stars) and output the accumulated intensity (shown by the brightness of the circle) during the exposure time of 3 ms (red line). An image corresponds to the intensity distribution according to the spatial arrangement of the photoreceptors, and a video is generated by acquiring images every 33 ms according to the predetermined frame rate $f = 30$ Hz. Note that the visual information during the 30 ms interval (white line) is completely lost. **b,** Visual representation by vidar. The photosensitive units (three circles) continuously capture photons and generate a spike (white circles) when the accumulated intensity exceeds a given threshold (here, the threshold is four photons). Vidar is a bit sequence array arranged in accordance with the spatial arrangement of the device, where a bit 1 indicates that a spike is generated by the photosensitive unit at that moment, and a bit

0 indicates that the unit is in the cumulative state. The vidar for the three photosensitive units here is $\begin{bmatrix} 01010001010100 \\ 00000001000000 \\ 01101010101011 \end{bmatrix}$.

The synchronous exposures and the same exposure time in images and videos impede the ability of digital cameras to capture the rapidly changing photonic world. However, such a design is not necessary. Here, we introduce vidar, a new visual representation that can better capture the temporal domain changes in light by utilizing a new temporal domain sampling mechanism and allowing asynchronous exposures. Vidar is the name of a god in Norse mythology. Here, vidar, as the combination of "vi-" (visual) and "-dar" (radar), without any capitalization, is coined to define a new form of visual information to replace video.

Specifically, vidar is no different from traditional images and videos in terms of spatial sampling. Vidar uses the same photosensitive devices, i.e., well-known CMOS or CCD sensors, as in traditional cameras. Therefore, vidar also takes advantage of the spatial arrangement of a lattice to represent spatial information (Fig. 2b). The fundamental difference between vidar and video is the use of a new temporal domain sampling method. All the photosensitive units continuously capture photons instead of being synchronously exposed with the same exposure time. When the accumulated intensity exceeds a given threshold, a spike is generated (Fig. 2b). The spike and the duration required to generate this spike are called a vit. The vits generated by each photosensitive unit are arranged in sequence according to chronological order. The simplest representation of vits is a bit stream, where 1 indicates that a spike appears at that moment, and 0 indicates that the unit is in the cumulative state. A bit 1 and all the 0s between this 1 and the previous 1 constitute a digital vit. Each photosensitive unit can generate a spike stream, and the spike streams generated by all the photosensitive units are arranged in accordance with the spatial arrangement of the device to form a spike stream array, that is, vidar.

The outstanding advantage of vidar over video is that the temporal domain change in light at each sampling position is effectively retained. With a device that is sensitive to a single photon, a photon can excite a spike. In this case, vidar records the exact and complete physical process. Ordinary photosensitive devices excite a spike only when they capture a set of photons, which is a rough representation of the physical process. However, the time relationship of the physical process is still preserved to the greatest extent, in contrast to when the time relationship is arranged uniformly at tens of hertz through artificial rules such as in video. In fact, the time sensitivity of CMOS photosensitive devices widely used today has reached tens of nanoseconds. With this new model vidar, high-speed temporal domain sampling of 10 million hertz can be achieved, and extremely fast physical processes can be recorded. Of course, daily vision applications do not require such a high sampling frequency. The first chip we developed set a sampling frequency of 40,000 Hz, which is 1,000 times faster than the sampling frequency of human vision and traditional cameras. Clearly, this chip can be utilized to shoot a high-speed rail with a speed of 350 km/h and a hard drive rotating at a speed of 7200 rpm.

Vidar records fine changes in light at various positions within a certain spatial range, and its physical meaning is very clear. Therefore, it is expected that vidar can be used to generate traditional images and videos. In fact, for any given moment, the scene radiance at each position and the pixel value of each pixel can be estimated from the vit covering that moment, and more detailed scene radiance and pixel value can be estimated by referring to the previous and spatially adjacent vits, thereby obtaining fine images at arbitrary moments (see Fig. 4 and the section on visual texture reconstruction with the vidar camera for details). The ability of vidar to reconstruct the scene radiance at any moment is called full-time imaging or continuous imaging.

*3.2  VidarOne chip and vidar camera system*

The VidarOne chip is developed based on the new visual representation model vidar and adopts an asynchronous pixel trigger architecture. As shown in Fig. 3a, the 400×250 pixel array converts the input photons into a spike stream array and utilizes a rolling shutter to detect the responses of all the pixels. After that, the row scanner scans the pixel array row by row. When one row of pixels is selected through the logic control signal, the data are transferred into the digital buffer for parallel readout. To support the high-speed output of the spike stream, the VidarOne chip provides an 8-channel specialized communication interface with a bandwidth of 500 megabits (Mb) per second. The synchronous readout interface is clocked at 20 MHz.

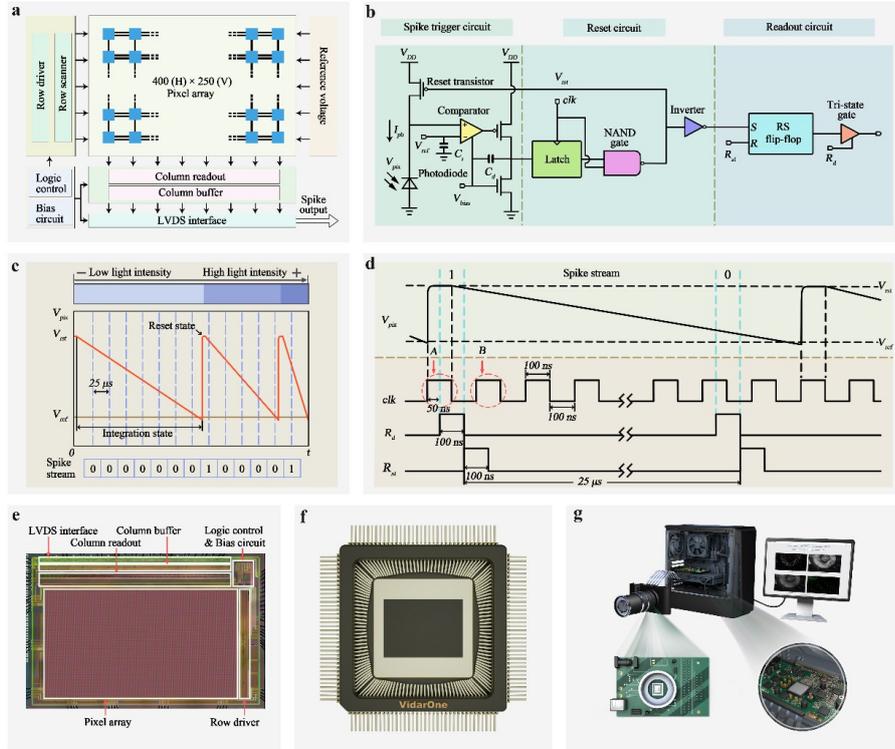

**Fig. 3. Design of the VidarOne chip and the vidar camera system. a**, Schematic of the chip architecture. It mainly consists of a pixel array, a row scanner with the configured driver, a column readout circuit with an addressable digital buffer, bias/reference circuits, and digital logic control. **b**, The in-pixel circuit includes three parts: a spike trigger circuit consisting of a photodiode, a reset transistor, and a comparator; a reset circuit consisting of a latch, a NAND gate and an inverter; and a readout circuit consisting of an RS flip-flop and a tri-state gate. $V_{pix}$ and $V_{ref}$ are the photodiode voltage and the reference (threshold) voltage, respectively; $V_{DD}$ and $V_{bias}$ are the supply voltage and the bias voltage, respectively; $I_{ph}$ is the photocurrent; $C_r$ and $C_d$ are the capacitors used to suppress the voltage fluctuation; $R_{st}$ is the row reset signal; $R_d$ is the row readout signal; and *clk* represents the clock signal, where the latch will be locked when *clk* is at a high level. **c**, Principle of spike triggering and spike coding. $V_{rst}$ is the reset state. The pixel intensity is encoded as 1 at the time when the flip signal (spike) is triggered and 0 otherwise. **d**, Timing diagram of the spike stream. **e**, Microphotograph of the VidarOne chip. Each fabricated block corresponds to the components in **a**. **f**, Image of the packaged VidarOne chip. **g**, The vidar camera system includes a visual information acquisition module composed of an industrial camera lens and a VidarOne chip, a high-speed sensing module implemented by a field-programmable gate array (FPGA) chip, and a real-time visual computing module implemented by a desktop workstation.

The basic circuit in a pixel, as shown in Fig. 3b, consists of a spike trigger circuit, a reset circuit and a readout circuit. The photodiode in the pixel continuously captures photons and converts the incident light illumination into a continuous photocurrent $I_{ph}$. Thereafter, the photodiode voltage $V_{pix}$ decreases during the collection of photoelectrons. When the photodiode voltage $V_{pix}$ reaches a certain threshold $V_{ref}$, the output of the comparator toggles, and a flip (spike) signal is generated (see Fig. 3c for illustration). The latch synchronizes the flip signal of the comparator under the enable operation of the clock signal $clk$. Once the latch detects the flip signal, the photodiode voltage $V_{pix}$ is reset to a predefined reset voltage. Meanwhile, the spike signal is sent to the RS flip-flop and saved. The row readout signal $R_d$ controls the sequential scanning and readout of the spike streams, and the row reset signal $R_{st}$ is responsible for clearing the signals in the RS flip-flop. The timing diagram of the spike pixel is shown in Fig. 3d. Under the control of the clock signal $clk$, the spike signal is fixed at a high level lasting 100 ns. For the spike signal generated at time A (see arrow A), the spike is read out after 50 ns by the row readout signal $R_d$ at a high level, considering that the row reset signal $R_{st}$ is at a low level. For the spike signal generated at time B (see arrow B), the RS flip-flop captures the spike signal after 50 ns, as the RS flip-flop is shielded through the reset signal $R_{st}$ at a high level. The spike is read out when the next readout signal $R_d$ arrives. In a readout cycle, only one spike signal can be processed even if two or more spike signals are triggered. The reason is that the RS flip-flop will not respond to the other spikes when it is latched by a spike signal.

As the row scanning time is 100 ns, the time resolution of spike streams generated by the 250-row pixel array is 25 μs.

Fabricated using standard 110-nm 1-poly 3-metal process technology, the VidarOne chip occupies a die area of 9.96×7.1 mm2 (Fig. 3e). Each square pixel has a size of 20×20 μm2 and achieves a 13.75% fill factor on the prototype chip. The large-size pixel detector can guarantee a sufficient photodetector area after placement of the metal grid. Under natural light, the chip can provide a high dynamic range of more than 100 dB without using a dynamic range enhancement technique. The energy consumption of the proposed design is approximately 370 mW. A physical view of the packaged VidarOne chip is shown in Fig. 3f. The placement and routing are carefully designed to minimize the silicon area for the pixel circuits.

The vidar camera system, which is composed of a visual information acquisition module, a high-speed sensing module, and a real-time visual computing module, is developed (Fig. 3g). The visual information acquisition module converts the input scene into spike streams, which are passed through the sensing module to undergo high-throughput real-time data processing operations and then are sent to the visual computing module through the peripheral component interconnect express (PCIE) bus.

*3.3 Visual texture reconstruction with the vidar camera*

The vidar camera has the ability of full-time imaging. The visual textures at any given moment can be reconstructed according to the characteristics of output spike streams (vits), and the dynamic range and quality of reconstructed textures are very flexible. To reconstruct the captured scene and bridge the gap between vidar data and conventional frame-based vision, we propose two visual texture reconstruction strategies, namely, texture from window (TFW) and texture from interspike interval (TFI) (see Methods for details).

Specifically, the TFW method takes advantage of the principle that the scene radiance is directly proportional to the spike count (firing rate); thus, one can compute the pixel value (proportional to the scene radiance) by using a moving time window to collect the spikes in a specific period (Fig. 4a). The reconstruction results are illustrated in Fig. 4c, where we present a novel vidar dataset called the vidar high-speed dataset (VHSSD) (see Methods). The first row of Fig. 4c presents the raw data of the vidar camera for eight different scenes, and the second row shows the texture reconstruction with TFW. The TFW method is suitable for stationary scenes. In the case of high-speed moving scenes, the scene radiance received by the vidar camera changes rapidly. At this time, the firing rate over a period cannot capture this rapid change in the scene radiance, causing blurry imaging (Fig. 4c second row). The TFI method is proposed to solve this problem by utilizing the fact that the scene radiance is inversely proportional to the interspike interval (Fig. 4b). Thus, only two spikes, i.e., one interspike interval, are needed to estimate the scene radiance in this period, which can match the rapid change in the scene radiance for high-speed moving scenes. In fact, the texture reconstructed with TFI updates the motion nearly synchronously. TFI achieves better results than TFW for high-speed moving scenes (Fig. 4c third row). We also compare our construction results quantitatively with that of conventional cameras. As illustrated in Table 1, our reconstruction methods achieve better results than conventional camera.

To facilitate the demonstration of the new idea, we develop a vidar camera simulator, Vidar-Sim, that can simulate arbitrary camera motion and object motion in 3D scenes and generate reliable spike streams similar to the vidar camera (see Methods). In addition, this simulator provides color images by simulating the RGB channels of the pixel. Here, we construct the scenes of "PKU flying ball" and "PKU coin" with Blender and generate spike streams with Vidar-Sim. The first and second rows of Fig. 4d present the reference images generated by an ordinary camera (frame rateHz) for the two scenes and the simulated spikes generated by Vidar-Sim, respectively. The TFW and TFI reconstruction results are shown in the third and fourth rows. The details in the reference images are fuzzy, while the images reconstructed based on the spike streams generated by Vidar-Sim show more texture details.

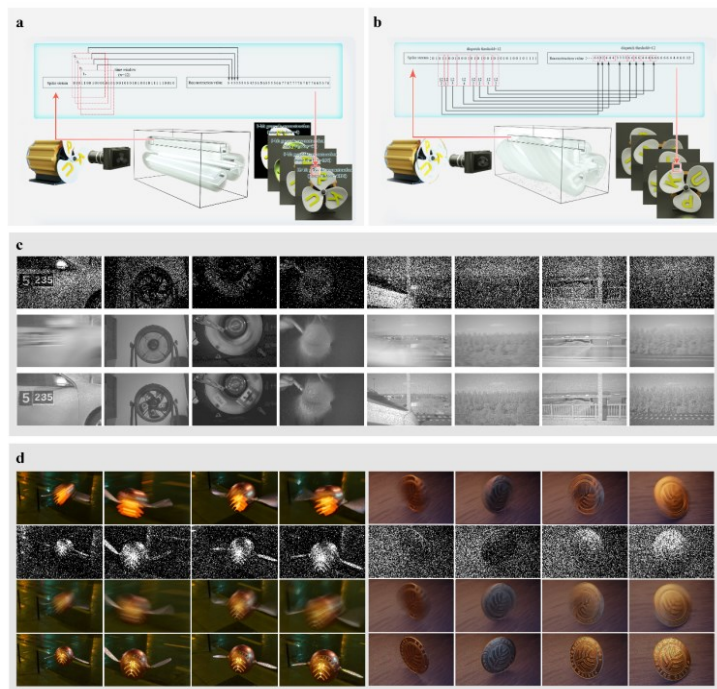

**Fig. 4. Texture reconstruction with the vidar camera**. **a**, Illustration of the TFW method. This method uses the principle that the scene radiance is directly proportional to the spike count. The light blue rectangle represents the spike streams of a pixel and the corresponding reconstructed grayscale value. The TFW method can reconstruct the texture with a free dynamic range by resizing the time window to different widths and collecting different numbers of spikes (see the four frames on the right). **b**, Illustration of the TFI method. This method uses the principle that the scene radiance is inversely proportional to the interspike interval. The TFI method applies to high-speed moving scenes. **c**, Reconstruction results for the VHSSD. The three rows represent the raw spikes from the vidar camera, texture reconstruction by TFW, and texture reconstruction by TFI. **d,** Reconstruction results for two scenes constructed with a Blender. Scene 1: PKU flying ball (inspired by the 'golden snitch' Quidditch game ball from the Harry Potter Series). The wings of the flying ball flap at night and enter the field of view of the vidar camera from far to near. Scene 2, PKU coin. A gold coin with the PKU logo rotates on a wooden desktop and eventually stops. Both scenes consider the slight movement of the camera.

*3.4 Super vision system with SNNs*

Here, we show that super vision can be achieved by combining the speed of the machine and the mechanism of biological vision. We propose a super vision system for high-speed moving object detection, tracking, prediction, and recognition based on SNNs that is 1000 times faster than the human vision system (Fig. 5a). Realizing these functions involves three major challenges: first, removing spikes generated from the background/static part of the scene for subsequent high-level visual tasks; second, detecting and smoothly tracking high-speed moving objects as well as predicting the trajectory; and third, recognizing the tracked objects. To accomplish these tasks, we propose a dynamic connection gate with short-term plasticity for filtering spatiotemporal spike sequences, a locally connected SNN for object detection and tracking, a continuous attractor neural network (CANN) for prediction, and a three-layer fully connected SNN for object recognition.

The detailed structure is shown in Fig. 5b-e. As the vidar camera generates spike streams with a fixed frequency for the background/static part of the scene, which will hamper the subsequent high-level visual tasks, the dynamic connection gate based on short-term plasticity is introduced here to filter spikes (Fig. 5b, see Methods). The gate closes when the input spike streams have a fixed frequency (corresponding to background or static objects) and opens when the spike frequency changes (corresponding to moving objects); thus, only the spike streams generated by the moving objects are retained. The neurons in the filter layer send excitatory postsynaptic potentials (EPSPs) to spatially adjacent neurons in the detection layer, where all the neurons fire spikes according to the leaky integrate-and-fire (LIF) model (Fig. 5d). Each moving object is found by detecting the connected area of the firing neurons, while in the tracking layer, different moving objects are associated by comparing the position or topology similarity of the moving neurons at the previous time with that at the current time. The next layer is a CANN (Fig. 5c), which is used to predict the trajectory by adding negative feedback to the neuronal dynamics. It can track a moving object anticipatively with an approximately constant leading time (see Methods). The recognition network is a multilayer fully connected SNN (Fig. 5e). The network is trained with the backpropagation spiking-timing-dependent plasticity (BP-STDP) learning rule (see Methods), and the recognition result is determined by the firing rate of the neurons in the last layer.

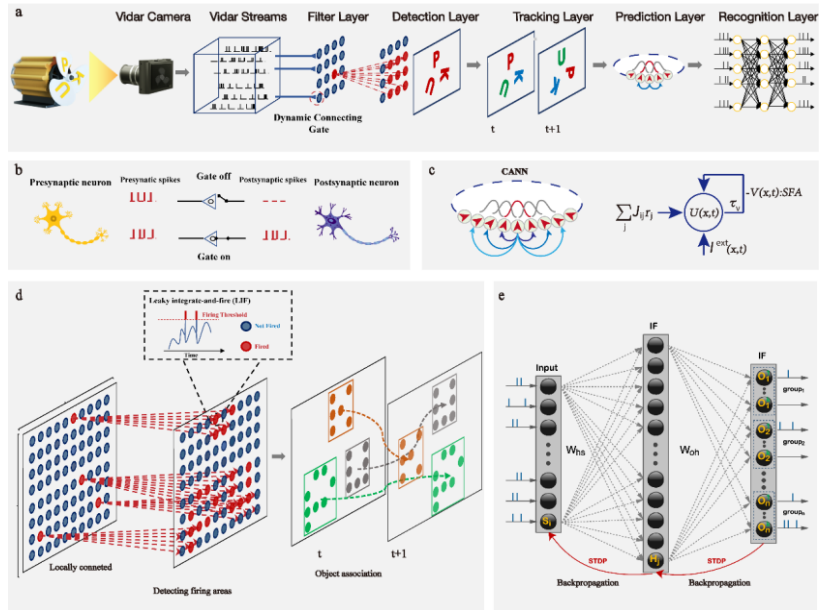

**Fig. 5. Super vision system. a,** Framework of high-speed moving object detection, tracking, prediction and recognition based on SNNs. **b,** Dynamic connection gate with short-term plasticity for the spatiotemporal spike sequence filter. It removes the spike streams that have a fixed frequency. **c,** CANN for predicting the trajectory. **d,** Locally connected SNN for object detection and tracking. **e,** Three-layer fully connected SNN trained for object recognition.

### 3.5 Demonstration of the utility of the vidar camera and the super vision system

To demonstrate the utility of the vidar camera and the super vision system, we design experiments of auxiliary referee and target pointing systems. Fig. 6a illustrates the auxiliary referee scene, in which we make use of a table tennis ball machine to launch a ball to simulate games such as tennis and badminton. The problem is to determine whether the ball is in or out of bounds (white line). As making a judgment by the human eye when the ball drop location is near the bounds is difficult, an eagle eye system is often used in the game. In addition to being expensive, the eagle eye system cannot record the moment when the ball hits the ground. Generally, it is calculated from the movement trajectory, which may cause disputes with a referee. In contrast, the vidar camera has the full-time imaging ability to record the entire process of ball landing, thus enabling the referee to determine the ball drop location (Fig. 6b).

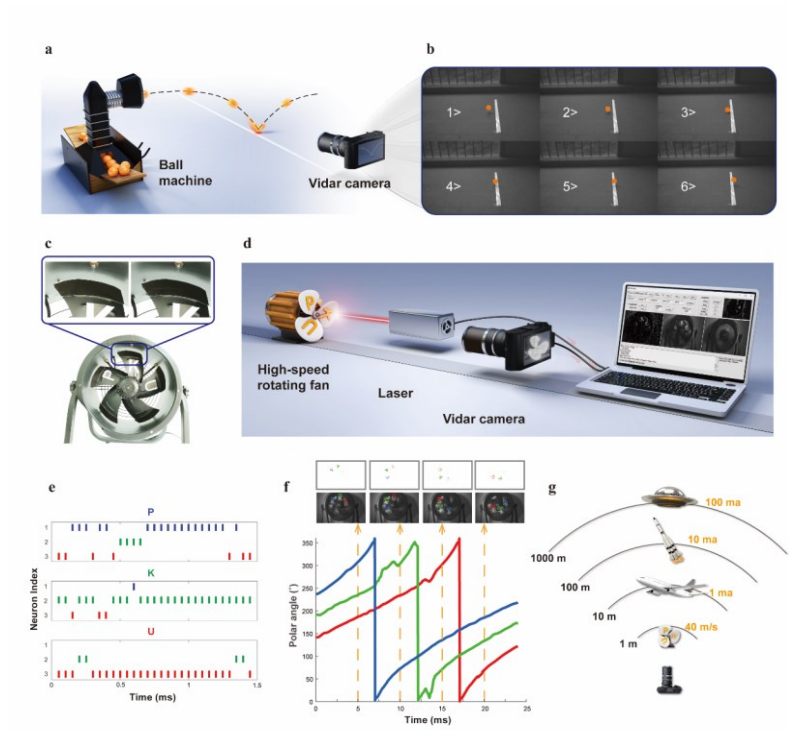

**Fig. 6. Demonstration of the utility of the vidar camera and the supervision system with auxiliary referee and target pointing systems. a,** Illustration of the auxiliary referee task, where the vidar camera is used to determine whether the ball is in or out of bounds. **b,** The vidar camera can capture the entire process of the ball landing. Here, the speed of the table tennis ball is approximately 100 km/h. Note that the ball and boundary are colored for emphasis, and only 6 of 170 frames are shown here. **c,** Comparison of a fan before and after laser hits. The laser sends 64 pulses and hits the predetermined character "K". **d,** Illustration of the target pointing system. The laser needs to hit the laser printing paper at the top of the known character on a high-speed rotating fan. **e,** SNN recognition test. The neuron in the output layer of the recognition SNN corresponding to the correct category produces the most spikes. **f,**

Multiobject tracking by detection. The y axis shows the polar angle of each object's position center point relative to the center of the fan. The SNNs can obtain a mask of each character and return its bounding box in real time. The mask and bounding box are colored by object membership. **g,** Evaluation of the performance of the vidar camera and the super vision system.

Moreover, a target pointing system is presented to demonstrate that high-speed vision can be achieved by combining the vidar camera and the super vision system (Fig. 6d). The vidar camera is set in front of a high-speed rotating fan (approximately 2400 rpm) with three characters ('P', 'K' and 'U') pasted on its blades. One can choose one of the characters as the pointing target, and the laser needs to fire and hit the photographic paper at the top of the character. Solving this problem requires three steps: first, detecting and tracking all the moving objects in the scene; second, recognizing all the moving objects and determining the position of the character given in advance; and third, predicting the trajectory of the character and controlling the laser to hit the target. The proposed super vision system (Fig. 5) is used to accomplish this task. Fig. 6e visualizes the output spike train in response to different characters. Fig. 6f illustrates the detection and tracking performance for the three characters, from which one can find that the network can detect all the moving objects and track them smoothly. Fig. 6c illustrates a comparison of the fan before and after laser hits. This demonstration provides an appropriate method to evaluate the performance of the vidar camera and the super vision system. The vidar camera and the super vision system can detect, track and recognize the fan moving with a linear velocity of 30 m/s within 0.75 m in real time (see Methods). According to the central perspective principle, this system can detect, track and recognize an aircraft flying at the speed of sound within 10 m. Moreover, it can detect, track and recognize a high-speed moving object at Mach 100 within 1 km (Fig. 6g).

## 3.6 Application prospects

The vidar camera is capable of capturing fast object movements. This camera has a high-speed mode with functions similar to the human eye and better performance than the human eye. This is not possible with traditional frame-based cameras due to the significant information loss between frames. By increasing the frame rate, some high-speed cameras, such as Phantom cameras, have mitigated this problem, but they require specialized sensors and shutters that are highly expensive. In contrast, the vidar camera is far more cost effective because it is built from conventional CCDs via regular semiconductor manufacturing processes. Thus, it can be widely used in daily life, such as with mobile phones and cameras.

Another advantage of the vidar camera over traditional cameras is that it offers a more flexible image acquisition method. The vidar camera can reconstruct the image at any given moment with considerable flexibility in the dynamic range. Distinct from another retina-inspired camera, the dynamic vision sensor (DVS) [20-22], whose photosensitive units only generate events when the brightness change exceeds a threshold, each photosensitive unit of the vidar camera keeps capturing photons independently and generates a spike when the accumulated intensity exceeds the given threshold. Therefore, the scene radiance at each sampling position is effectively recorded by the vidar camera. We believe that this system will create its own niche in surveillance systems, with applications to dynamic face recognition, fingerprint recognition, and palm print recognition.

The vidar camera is inspired by the neuronal circuitry structure and information processing mechanism in the primate fovea. It converts light signals into electrical signals, yielding spike trains as output that can be naturally processed by SNNs. Given that SNNs are efficient and effective in visual perception and cognitive tasks, we expect that the combination of the vidar camera and the super vision system based on SNNs will provide plentiful utilities for both fundamental research questions and practical applications, such as object detection, recognition and tracking at electrical speeds.

## 4. Conclusion

By replacing video with vidar, vidar cameras will bring camera development back on the right track, realizing the technological potential of optoelectronic technology that has been suppressed for decades and replacing traditional video cameras in almost all fields, especially for high-speed scenes, thus triggering a revolution in the camera field.

The essence of vidar is a spike stream that characterizes the process of optical temporal and spatial changes, which is a natural input of SNNs. Vidar, as a new generation of the eye of machine vision, will play an important role in the era of artificial intelligence.

## Acknowledgements

This work was supported by projects of the National Natural Science Foundation of China (61425025) and the Beijing Municipal Science and Technology Project (Z151100000915070, Z171100000117008).